\documentclass[runningheads]{llncs}

\usepackage[firstpage]{draftwatermark} 

\usepackage[T1]{fontenc}
\usepackage{graphicx}
\usepackage{marvosym}

\usepackage{numprint}
\usepackage{paralist}

\usepackage{amsmath}
\usepackage{amssymb}
\usepackage{xifthen}

\usepackage{algorithm}
\usepackage{algpseudocode}
\usepackage{algorithmicx}

\usepackage{cite}
\usepackage{comment}

\usepackage{pgfplots}
\usepackage{tikz}
\usetikzlibrary{positioning, matrix, quotes, calc, shapes.multipart, fit, patterns}

\usepackage[frozencache,cachedir=minted_cache]{minted2}

\usepackage{array}
\usepackage{subcaption}
\usepackage{hyperref}

\usepackage{mathtools}

\newcommand{\set}[1]{\left\{#1\right\}}

\newcommand{\seq}[1]{\langle #1 \rangle}
\newcommand{\tuple}[1]{\seq{#1}}

\newcommand{\N}[0]{\mathbb{N}}

\tikzset{every node/.style={node distance=1em, align=center}}
\tikzset{state/.style={draw, circle, inner sep=2pt}}

\newcommand{\fallback}[2]{\ifthenelse{\isempty{#1}}{#2}{#1}}

\tikzset{super state/.style={shape=rectangle, draw, node contents=}}
\tikzset{trans/.style={draw,-latex}}
\tikzset{point/.style={inner sep=0, node contents=}}

\newcommand{\transIOA}[1][]{\tuple{q_{#1},i_{#1},o_{#1},q'_{#1}}}
\newcommand{\IOA}[0]{\tuple{Q,q_0,I,O,T}}

\newcommand{\model}[0]{\mathcal{M}}

\newcommand{\code}[1]{\texttt{#1}}
\newminted{python}{mathescape,linenos,numbersep=5pt,autogobble,fontsize=\small,frame=single, escapeinside=||}

\SetWatermarkAngle{0}
\SetWatermarkText{\raisebox{12.5cm}{
\hspace{-0.25cm}
\href{https://doi.org/10.5281/zenodo.15296685}{\includegraphics{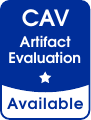}}
\hspace{8.3cm}
\includegraphics{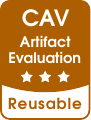}
}}

\begin{document}

\title{Extending AALpy with Passive Learning:\\ A Generalized State-Merging Approach}
\titlerunning{Extending AALpy with Passive Learning}

\newcommand{\correspondingAuthor}[0]{\textsuperscript{(\Letter)}}
\newcommand{\authorDef}[4][]{#2\inst{#3}#1\orcidID{#4}}

\newcommand{\BenjaminVonBerg}[2][]{\authorDef[#1]{Benjamin von Berg}{#2}{0009-0001-3595-4715}}
\newcommand{\BernhardAichernig}[2][]{\authorDef[#1]{Bernhard K. Aichernig}{#2}{0000-0002-3484-5584}}
\newcommand{\DarkoStern}[2][]{\authorDef[#1]{Darko \v{S}tern}{#2}{0000-0003-3449-5497}}
\newcommand{\MartinTappler}[2][]{\authorDef[#1]{Martin Tappler}{#2}{0000-0002-4193-5609}}

\author{
  \BenjaminVonBerg[\correspondingAuthor]{1} \and\\
  \BernhardAichernig{1,2}
}


\newcommand{\SAI}[1]{
  Institute of Software Engineering and Artificial Intelligence, \\ 
  Graz University of Technology, Austria \\ 
  \email{#1@tugraz.at}
}

\newcommand{\JKU}[1]{
    Institute for Formal Models and Verification,\\ 
    Johannes Kepler University Linz, Austria \\
    \email{#1@jku.at}
}

\institute{
  \SAI{benjamin.vonberg} \and
  \JKU{bernhard.aichernig}
}

\maketitle

\begin{abstract}
AALpy is a well-established open-source automata learning library written in Python with a focus on active learning of systems with IO behavior. It provides a wide range of state-of-the-art algorithms for different automaton types ranging from fully deterministic to probabilistic automata.
In this work, we present the recent addition of a generalized implementation of an important method from the domain of passive automata learning: state-merging in the red-blue framework. Using a common internal representation for different automaton types allows for a general and highly configurable implementation of the red-blue framework. We describe how to define and execute state-merging algorithms using AALpy, which reduces the implementation effort for state-merging algorithms mainly to the definition of compatibility criteria and scoring. This aids the implementation of both existing and novel algorithms. In particular, defining some existing state-merging algorithms from the literature with AALpy only takes a few lines of code.

 \and red-blue framework
\end{abstract}

\section{Introduction}

AALpy \cite{MuskardinAPPT22} is an active automata learning library written in Python with a focus on systems that show IO behavior, i.e. reactive systems. Automata learning algorithms extract a finite-state model from observations of a system-under-learning. They form the basis for black-box checking \cite{peled1999black,DBLP:journals/fmsd/AichernigT19}. 

AALpy is open-source and published on github\footnote{\url{https://github.com/DES-Lab/AALpy}}. It is well-established in the community and has been used in various domains. This includes reinforcement learning \cite{TapplerPAK24, TapplerPKMBL22}, analysis of communication protocols \cite{PferscherA21, PferscherA22} and neural networks \cite{MuskardinAPT22}, black-box checking \cite{ShijuboWS23, KuzeSU23}, anomaly detection \cite{ModdemannSDPN24}, software testing \cite{Ganty24}, process mining \cite{BolligFN24,KobialkaPBJT24} and reuse in other learning algorithms \cite{AyoughiNSS24, JungesR22}. 

Passive learning algorithms use existing training data, like system logs, while active learning queries the system during learning. So far, AALpy has been mainly concerned with active automata learning.
In this paper, we extend AALpy with a passive automata learning library for state-merging algorithms based on the red-blue framework \cite{LangPP98}. 
This encompasses (1) algorithms already implemented in AALpy, such as RPNI \cite{RPNI} and IOAlergia \cite{IOAlergia}, (2) existing algorithms that were not originally designed for the IO setting, such as the likelihood-ratio method by Verwer et al. \cite{VerwerWW10} or EDSM \cite{LangPP98} and (3) new approaches based on state-merging, be it completely novel methods or adaptations of existing methods in order to account for domain knowledge, such as additional constraints.

The rest of the paper is structured as follows. Section~\ref{sec:pop} gives some background and discusses how we generalized state merging. Section~\ref{sec:usage} demonstrates the usage of the new library. Section~\ref{sec:conc} draws the conclusions, relates AALpy to other tools and gives an outlook of further possible extensions. Further examples are provided in the \hyperref[appendix]{Appendix}.

\section{Principles of Operation}
\label{sec:pop}

\subsection{Universal Internal Formalism}

The tool presented in this work covers several types of finite-state automata with IO behavior. These automaton types share a small set of common assumptions: 
(1) There is a unique initial state from which execution starts.
(2) Applying an input to an automaton triggers a state change and the emission of an output.
(3) For each transition, the reached state is uniquely determined by the combination of current state, applied input and observed output.
The following definition of automata covers Assumption 1 and 2 above and can be extended to more specific types of automata.

\begin{definition}
    \label{def:IOA}
    An \textbf{IO automaton} is a tuple $\model = \tuple{Q, q_0, I, O, T}$, where $Q$ is the set of states, $q_0 \in Q$ is the initial state, $I$ and $O$ are the sets of input and output symbols respectively and $T \subseteq Q \times I \times O \times Q$ is a set of transitions.
\end{definition}

A transition $\transIOA \in T$ implies that applying input $i$ in state $q$ may cause $\model$ to emit the output $o$ and switch to state $q'$. In other words, the automaton might behave nondeterministically. We discern between different forms of determinism.

\begin{definition}\label{def:Determinism}
    An automaton $\model = \IOA$ is \textbf{deterministic} iff
    \[
        \forall \transIOA[1], \transIOA[2] \in T : q_1 = q_2 \land i_1 = i_2 \Rightarrow o_1 = o_2 \land q'_1 = q_2'
    \]
    It is \textbf{observably nondeterministic} iff
    \[
        \forall \transIOA[1], \transIOA[2] \in T : q_1 = q_2 \land i_1 = i_2 \land  o_1 = o_2 \Rightarrow q'_1 = q_2'
    \]
\end{definition}

Note that observable nondeterminism is equivalent to Assumption 3 above. Deterministic and observably nondeterministic automata allow writing the set of transitions as a (partial) transition function $\delta$ of the set of transitions with $\delta : Q \times I \rightharpoonup O \times Q$ and $\delta : Q \times I \times O \rightharpoonup Q$ respectively. 

The definition of IO automata follows the behavior of Mealy machines, where the output of a transition depends on the source state, the target state and the provided input symbol. This is in contrast to Moore behavior, where the output depends on the target state alone:
\begin{definition}
    \label{def:Moore property}
    An IO automaton $\model = \IOA$ has \textbf{Moore behavior} iff
    \[
        \forall \transIOA[1], \transIOA[2] \in T: q'_1 = q'_2 \Rightarrow o_1 = o_2
    \]
\end{definition}

A Moore machine can thus be seen as a deterministic IO automaton $\model = \IOA$ with Moore behavior that emits the output $o_0$ associated with $q_0$ prior to handling the first input. Informally, we treat the emission of the initial output as the result of a transition $\tuple{\epsilon, \epsilon, o_0, q_0} \in T$. 

When learning probabilistic automaton types, not only the possibility of a transition happening needs to be considered, but also its probability. This prompts the use of IO Frequency Automata (IOFA), which augment IOA with an observation count for transitions.

\begin{definition}
    \label{def:IOFA}
    An \textbf{IO frequency automaton} is a tuple $\model = \tuple{Q, q_0, I, O, \delta, \nu}$, where $\tuple{Q, q_0, I, O, \delta}$ is an observably nondeterministic IO automaton and $\nu : Q \times I \times O \rightarrow \N$ is the frequency function.
\end{definition}

The intuition behind $\nu(q,i,o) = n$ is that the transition from $q$ on input $i$ emitting $o$ has been observed $n$ times.
We use IOFA as the common internal representation from which all supported automaton types can be extracted. This allows us to use a single implementation which is shared across all supported automaton types. Additional restrictions are required during the merging process to enforce the structural form of those automaton types and are covered in Section \ref{sec:enforce structure}.

\subsection{State Merging in the Red-Blue Framework}

This section, gives a brief overview over state merging algorithms and the red-blue framework. For a more detailed account consider \cite{de2010grammatical} and \cite{LangPP98} respectively.

State-merging algorithms are one of the main categories of passive automata learning algorithms. The input is given as a collection of traces, where each trace is a sequence of input-output pairs. The basic structure of state-merging algorithms consists of the following steps. First, a simple automaton representation of the provided traces is created, commonly using a so-called Prefix Tree Automaton (PTA). The PTA contains a state for every prefix in the input data. Grouping the behavior of traces with shared prefixes is admissible because we assume observable nondeterminism and results in a tree-shaped IOFA that describes exactly the observed traces and no other behavior. In the main loop of the algorithm, the automaton is generalized by iteratively merging pairs of states until no more merges are possible. We use the term \emph{merge candidate} to refer to a pair of states that is considered for merging. 
Merging two states $q_1$ and $q_2$ results in a state that exhibits the behavior of both states. 
If both $q_1$ and $q_2$ have a transition with input $i$ and output $o$ leading to states $q'_1$ and $q'_2$ respectively, the assumption of observable nondeterminism implies that $q'_1$ and $q'_2$ must be the same state. 
Therefore, merging $q_1$ and $q_2$ also requires merging $q'_1$ and $q'_2$. This is called an \emph{implied merge}. Together, the initial merging of $q_1$ and $q_2$ and its implied merges results in a \emph{partitioning} of the state space into sets of (transitively) merged states. The criteria for selecting potential merge candidates and evaluating their compatibility and quality are specific to the used algorithm.

\begin{example}[PTAs and Merging]\label{ex:PTA and Merging}
    The left side of Figure \ref{fig:PTA and Merge} shows the PTA for the three traces $\tuple{\tuple{x,a},\tuple{x,a},\tuple{x,a}}$, $\tuple{\tuple{x,a},\tuple{x,a},\tuple{y,b}}$ and $\tuple{\tuple{y,b}}$. Trying to merge $q_1$ and $q_2$ (dashed line) will lead to the implied merge of $q_2$ and $q_4$ (dotted line), which in turn requires merging not only $q_4$ and $q_5$ but also $q_3$ and $q_6$ since $q_1$ and $q_4$ are in the same partition. Applying the merge results in the automaton shown on the right where $p_1=\set{q_1,q_2,q_4,q_5}$ and $p_2=\set{q_3,q_6}$.
    
    \begin{figure}[t]
    \centering
    \begin{tikzpicture}
    \tikzset{merge/.style={draw,latex-latex, color=gray}}

    \node (q_0) {};
    \node[state, right=of q_0] (q_1) {$q_1$}; 
    \node[state, right=2em of q_1] (q_2) {$q_2$}; 
    \node[state, right=2em of q_2] (q_3) {$q_4$}; 
    \node[state, right=2em of q_3] (q_4) {$q_5$};
     
    \node[state, below=of q_2] (q_5) {$q_3$};
    \node[state, below=of q_4] (q_6) {$q_6$};

    \draw[trans] (q_0) edge (q_1); 
    \draw[trans] (q_1) edge node[above] {$x / a$} (q_2); 
    \draw[trans] (q_2) edge node[above] {$x / a$} (q_3); 
    \draw[trans] (q_3) edge node[above] {$x / a$} (q_4); 
    \draw[trans] (q_1) edge node[below left] {$y / b$} (q_5); 
    \draw[trans] (q_3) edge node[below left] {$y / b$} (q_6); 

    \draw[merge, dashed] (q_1) edge[bend angle=22, bend right] (q_2);
    \draw[merge, dotted] (q_2) edge[bend angle=22, bend right] (q_3);
    \draw[merge, dotted] (q_3) edge[bend angle=22, bend right] (q_4);
    \draw[merge, dotted] (q_5) edge (q_6);

    \node[right=2em of q_4] (p_0) {};
    \node[state, right=of p_0] (p_1) {$p_1$};
    \node[state, below=of p_1] (p_2) {$p_2$};

    \draw[trans] (p_0) edge (p_1); 
    \draw[trans] (p_1) edge[out=40, in=-40, looseness=4] node[right] {$x / a$} (p_1); 
    \draw[trans] (p_1) edge node[left] {$y / b$} (p_2); 
    
\end{tikzpicture}
    \caption{Example of a PTA without frequencies (left) and the partitioning resulting from merging states $q_1$ and $q_2$ (right).}\label{fig:PTA and Merge}
    \end{figure}
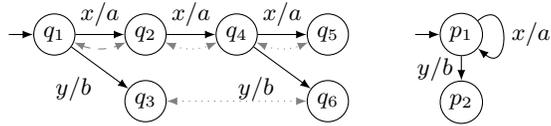
\end{example}

\begin{algorithm}[t]
    \caption{State Merging in the Red-Blue Framework.}

\textbf{Input:} Collection of traces $\mathcal{TR}$ \\
\textbf{Parameters:} Scoring function $score$, state order $<$ \\
\textbf{Output:} Automaton model $\model$ \\
\textbf{Algorithm:}
\begin{algorithmic}[1]
    \State $\model \gets$ PTA created from $\mathcal{TR}$
    \State $Q_r \gets \set{q_0^\model}$ \Comment{Initialize red states to initial state}
    \Loop
        \State $Q_b \gets $ set of blue states derived from $Q_r$
        \If {$Q_b = \emptyset$} 
            \State \Return $\model$ \Comment{Terminate if no more merge candidates are available} 
        \EndIf
        \State $S \gets \set{b \in Q_b\ |\ \forall r \in Q_r: score(\model, r, b) = -\infty}$ \Comment{Unmergeable blue states}
        \If {$S \neq \emptyset$}
            \State $Q_r \gets Q_r \cup \set{\min_< S}$ \Comment{Promote minimal unmergeable blue state}
        \Else 
            \State $\tuple{r,b} \gets \textrm{argmax}_{\tuple{r, b} \in Q_r \times Q_b} score(\model, r, b)$ \Comment{Select best merge wrt. $score()$}
            \State $\model \gets merge(\model, r, b)$ \Comment{Update model}
        \EndIf
    \EndLoop
\end{algorithmic}
    \label{alg:RedBlue}
\end{algorithm}

Algorithm \ref{alg:RedBlue} gives a high level description of the red-blue framework. The algorithm is parameterized by a scoring function $score$ and an order relation over states $<$. The red-blue framework restricts which pairs of states are considered for merging in each iteration of the algorithm in order to reduce the number of compatibility calculations. This is achieved by dynamically dividing the nodes into three categories throughout the execution of the algorithm. The set of red states contains states that are considered mutually distinct. It is initialized to consist only of the root node of the PTA. The set of blue states is defined as the set of all states reachable from red states in a single transition that are not themselves red. The set of white states consists of all states that are neither red nor blue. In each iteration of the algorithm, only pairs consisting of a red and a blue state are considered for merging. If a blue state is incompatible with all red states, it is promoted to a red state. Otherwise the merge candidate with the best performing score according to an algorithm-specific function is chosen as the next intermediate model. For both promotion and merging, if there are several eligible candidates, ties are broken using a specified node order\footnote{Omitted in line 12 for brevity.}. When all states are red, no more merges are possible and the final model is returned.

\begin{example}[Red-Blue Framework]\label{ex:RedBlue}
    Consider a slightly modified PTA compared to Example \ref{ex:PTA and Merging}, where the transition from $q_4$ to $q_6$ on input $y$ emits $a$ instead of $b$. Figure \ref{fig:RedBlue} shows the progression from the PTA (left) to the final automaton (right) when applying the red-blue framework for learning a deterministic Mealy machine from the provided data. Nodes are ordered according to the shortlex order of their prefixes and labeled $q_1$ to $q_6$ accordingly. States $q_1$ and $q_2$ can not be merged due to the transitions on input $y$ from states $q_1$ and $q_4$ emitting different outputs. The state $q_2$ is promoted since it is the minimal blue node and incompatible with the only red state $q_1$. Due to the order of nodes, $q_1$ and $q_3$ are merged before merging $q_2$ and $q_4$. As a result of the latter merge, $q_6$ becomes a child of $q_2$ and is finally merged with $q_1$.
    
    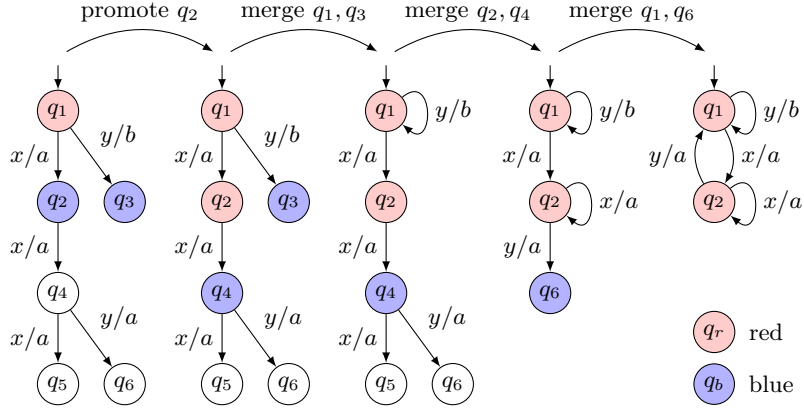
\begin{figure}[t]
    \centering
    \begin{tikzpicture}
    \newcommand{\FileDistance}[0]{6em}
    \tikzset{red/.style={state, fill=red!20}}
    \tikzset{blue/.style={state, fill=blue!30}}

    \node (a_0) {};
    \node[red, below=of a_0] (a_1) {$q_1$}; 
    \node[blue, below=2em of a_1] (a_2) {$q_2$}; 
    \node[state, below=2em of a_2] (a_4) {$q_4$}; 
    \node[state, below=2em of a_4] (a_5) {$q_5$};
     
    \node[blue, right=of a_2] (a_3) {$q_3$};
    \node[state, right=of a_5] (a_6) {$q_6$};

    \draw[trans] (a_0) edge (a_1); 
    \draw[trans] (a_1) edge node[left] {$x / a$} (a_2); 
    \draw[trans] (a_2) edge node[left] {$x / a$} (a_4); 
    \draw[trans] (a_4) edge node[left] {$x / a$} (a_5); 
    \draw[trans] (a_1) edge node[above right] {$y / b$} (a_3); 
    \draw[trans] (a_4) edge node[above right] {$y / a$} (a_6); 
    
    \node[right=\FileDistance of a_0] (b_0) {};
    \node[red, below=of b_0] (b_1) {$q_1$}; 
    \node[red, below=2em of b_1] (b_2) {$q_2$}; 
    \node[blue, below=2em of b_2] (b_4) {$q_4$}; 
    \node[state, below=2em of b_4] (b_5) {$q_5$};
     
    \node[blue, right=of b_2] (b_3) {$q_3$};
    \node[state, right=of b_5] (b_6) {$q_6$};

    \draw[trans] (b_0) edge (b_1); 
    \draw[trans] (b_1) edge node[left] {$x / a$} (b_2); 
    \draw[trans] (b_2) edge node[left] {$x / a$} (b_4); 
    \draw[trans] (b_4) edge node[left] {$x / a$} (b_5); 
    \draw[trans] (b_1) edge node[above right] {$y / b$} (b_3); 
    \draw[trans] (b_4) edge node[above right] {$y / a$} (b_6); 
    
    \node[right=\FileDistance of b_0] (c_0) {};
    \node[red, below=of c_0] (c_1) {$q_1$}; 
    \node[red, below=2em of c_1] (c_2) {$q_2$}; 
    \node[blue, below=2em of c_2] (c_4) {$q_4$}; 
    \node[state, below=2em of c_4] (c_5) {$q_5$};
     
    \node[state, right=of c_5] (c_6) {$q_6$};

    \draw[trans] (c_0) edge (c_1); 
    \draw[trans] (c_1) edge node[left] {$x / a$} (c_2); 
    \draw[trans] (c_2) edge node[left] {$x / a$} (c_4); 
    \draw[trans] (c_4) edge node[left] {$x / a$} (c_5); 
    \draw[trans] (c_1) edge[out=40, in=-40, looseness=4] node[right] {$y / b$} (c_1); 
    \draw[trans] (c_4) edge node[above right] {$y / a$} (c_6); 
    
    \node[right=\FileDistance of c_0] (d_0) {};
    \node[red, below=of d_0] (d_1) {$q_1$}; 
    \node[red, below=2em of d_1] (d_2) {$q_2$}; 
     
    \node[blue, below=2em of d_2] (d_6) {$q_6$};

    \draw[trans] (d_0) edge (d_1); 
    \draw[trans] (d_1) edge node[left] {$x / a$} (d_2); 
    \draw[trans] (d_1) edge[out=40, in=-40, looseness=4] node[right] {$y / b$} (d_1); 
    \draw[trans] (d_2) edge[out=40, in=-40, looseness=4] node[right] {$x / a$} (d_2); 
    \draw[trans] (d_2) edge node[left] {$y / a$} (d_6); 
    
    \node[right=\FileDistance of d_0] (e_0) {};
    \node[red, below=of e_0] (e_1) {$q_1$}; 
    \node[red, below=2em of e_1] (e_2) {$q_2$}; 

    \draw[trans] (e_0) edge (e_1); 
    \draw[trans] (e_1) edge[bend left] node[right] {$x / a$} (e_2); 
    \draw[trans] (e_2) edge[bend left] node[left] {$y / a$} (e_1); 
    \draw[trans] (e_1) edge[out=40, in=-40, looseness=4] node[right] {$y / b$} (e_1); 
    \draw[trans] (e_2) edge[out=40, in=-40, looseness=4] node[right] {$x / a$} (e_2); 

    \draw[-latex] (a_0) edge[bend left] node[above] {promote $q_2$} (b_0);
    \draw[-latex] (b_0) edge[bend left] node[above] {merge $q_1, q_3$} (c_0);
    \draw[-latex] (c_0) edge[bend left] node[above] {merge $q_2, q_4$} (d_0);
    \draw[-latex] (d_0) edge[bend left] node[above] {merge $q_1, q_6$} (e_0);

    \node[blue] at (c_6 -| e_2) (blue legend) {$q_b$};
    \node[red, above=1ex of blue legend] (red legend) {$q_r$};
    \node[right=0.5ex of blue legend] {blue};
    \node[right=0.5ex of red legend] {red};
    
\end{tikzpicture}
    \caption{Intermediate steps of state merging in the red-blue framework.}\label{fig:RedBlue}
    \end{figure}
\end{example}

\subsection{Enforcing Automaton Structure}\label{sec:enforce structure}

In this section, we describe how the automaton structure of the different supported automaton types is enforced in the internal representation.
We categorize the automaton types supported by AALpy in two different ways: (1) output behavior and (2) transition behavior. The output behavior of states may depend on both inputs and the current state, like in Mealy machines, or on states alone, like in Moore machines. Transition behavior may be deterministic, nondeterministic or probabilistic. Taking the product of these two categorization schemes (1 and 2) results in six different automaton types also supported in AALpy. 
AALpy also supports two additional types for DFAs and discrete time Markov chains. However, these two types are subsumed by the other types. DFAs can be seen as (deterministic) Moore machines with binary output, representing accept and reject. Markov chains have the same semantics as probabilistic Moore machines with a single input that represents the passage of time. Table \ref{tab:behavior-types} shows the different combination of behavior types and the respective classes in AALpy.

\begin{table}[t]
    \centering
    \caption{Combination of output behavior (rows) and transition behavior (columns) and the respective automaton classes in AALpy.}
    \label{tab:behavior-types}
    \begin{tabular}{l|c|c|c}
        & Deterministic & Nondeterministic & Probabilistic \\
        \hline
        Moore  & \code{MooreMachine}, \code{Dfa} & \code{NDMooreMachine} & \code{Mdp}, \code{MarkovChain} \\
        Mealy & \code{MealyMachine} & \code{Onfsm} & \code{StochasticMealyMachine} \\
    \end{tabular}
\end{table}

\textbf{Output Behavior.} Mealy output behavior is the default for IOFAs and thus does not need to be enforced. This is not the case for Moore behavior. However, since each PTA has only one predecessor, Moore behavior (Def. \ref{def:Moore property}) is ensured at the beginning of the state-merging algorithm. The Moore property can be turned into an invariant of state-merging by strengthening the local compatibility criterion during merging such that merges which lead to a violation of the Moore property are rejected. Consequently, Moore behavior of the final automaton is ensured.

\textbf{Transition Behavior.} If deterministic transition behavior is given for a PTA, we can enforce it in the final automaton analogously to Moore behavior by rejecting merge candidates that lead to nondeterministic behavior (Def. \ref{def:Determinism}).
Similar to Mealy behavior, nondeterministic transition behavior is the default for IOFA.
To extract an automaton with probabilistic transition behavior from an IOFA $\tuple{Q, q_0, I, O, \delta, \nu}$, the frequency function needs to be normalized. The probability distribution $p_{q,i}(o, q')$ describing the transitions originating from state $q$ on input $i$ to state $q'$ emitting output $o$ is given as
\[
p_{q,i}(o, q') = 
\begin{cases}
\nu(q,i,o) / \sum_{o' \in O} \nu(q,i,o')  & \mathrm{if}\ \delta(q,i,o) = q' \\
0 & \mathrm{otherwise}
\end{cases}
\]

\section{Usage}
\label{sec:usage}

\subsection{Defining and Running State-Merging Algorithms}

To define a state-merging algorithm, the \code{GeneralizedStateMerging} class is instantiated. Its most important parameters are:
\begin{description}
    \item \code{output\_behavior}: Defines whether an automaton with Moore or Mealy behavior should be learned and adds corresponding restrictions for local compatibility. Possible values are \code{"moore"} and \code{"mealy"}.
    \item \code{transition\_behavior}: Defines whether the learned automaton is assumed to be deterministic, nondeterministic or probabilistic and adds corresponding restrictions for local compatibility. The values are \code{"deterministic"}, \code{"nondeterministic"} and \code{"stochastic"} respectively.
    \item \code{score\_calc}: An object of type \code{ScoreCalculation}, which governs the details of local compatibility and score calculation.
\end{description}

The defined algorithm can be executed using the object's \code{run} method, which requires a single argument representing the collection of available traces. Each trace is expected to be a sequence of input-output pairs. Algorithms with Moore behavior also expect the traces to be prepended with the initial output. Other input formats are discussed below. Depending on the combination of the arguments \code{output\_behavior} and \code{transition\_behavior}, the \code{run} method determines what kind of automaton is extracted from the internal representation.

\begin{example}\label{ex:RPNI}
    \begin{figure}[t]
        \centering
        \newcommand{\FileDistance}[0]{6em}
\newcommand{\NodeDist}[0]{2em}

\newcommand{\NoAlarm}[0]{N}
\newcommand{\Alarm}[0]{A}
\newcommand{\Armed}[0]{\NoAlarm}
\newcommand{\UnArmed}[0]{\NoAlarm}
\newcommand{\Faulty}[0]{\NoAlarm}

\newcommand{\unlock}[0]{l}
\newcommand{\lock}[0]{l}
\newcommand{\open}[0]{d}
\newcommand{\close}[0]{d}

\begin{tabular}[c]{cc}
\subfloat[Correct car alarm system.]{
\begin{tikzpicture}
    \node (q_0) {};
    \node[state, right=of q_0] (q_3) {\Armed};
    \node[state, below=\NodeDist of q_3] (q_4) {\Alarm};
    \node[state, left=\NodeDist of q_4] (q_6) {\UnArmed};
    \node[state, below=\NodeDist of q_4] (q_2) {\UnArmed};
    \node[state, right=\NodeDist of q_4] (q_5) {\Alarm};
    \node[state, right=\NodeDist of q_5] (q_1) {\UnArmed};

    \draw[trans] (q_0) edge (q_3);
    \draw[trans] (q_3) edge[bend left] node[above right]{\unlock} (q_1);
    \draw[trans] (q_3) edge node[left]{\open} (q_4);
    \draw[trans] (q_1) edge node[above right]{\lock} (q_3);
    \draw[trans] (q_1) edge node[below right]{\open} (q_2);
    \draw[trans] (q_4) edge node[right]{\unlock} (q_2);
    \draw[trans] (q_4) edge[bend left] node[above]{\close} (q_5);
    \draw[trans] (q_5) edge[bend left] node[below]{\open} (q_4);
    \draw[trans] (q_5) edge node[above left]{\unlock} (q_1);
    \draw[trans] (q_2) edge[bend right] node[below right]{\close} (q_1);
    \draw[trans] (q_2) edge node[above right]{\lock} (q_6);
    \draw[trans] (q_6) edge[bend right] node[below left]{\unlock} (q_2);
    \draw[trans] (q_6) edge node[above left]{\close} (q_3);    
\end{tikzpicture}\label{fig:Det Car}
}

&

\subfloat[Faulty car alarm system.]{
\begin{tikzpicture}
    \node (q_0) {};
    \node[state, right=of q_0] (q_3) {\Armed};
    \node[state, right=\NodeDist of q_3] (q_5) {\Alarm};
    \node[state, above=\NodeDist of q_5] (q_1) {\UnArmed};
    \node[state, below=\NodeDist of q_5] (q_4) {\Alarm};
    \node[state, right=\NodeDist of q_5] (q_2) {\UnArmed};
    \node[state, right=\NodeDist of q_2] (q_6) {\UnArmed};

    \node[state, right=\NodeDist of q_6] (q_7) {\Faulty};

    \draw[trans] (q_0) edge (q_3);
    \draw[trans] (q_3) edge[bend left] node[above left]{\unlock} (q_1);
    \draw[trans] (q_3) edge[bend right] node[below left]{\open} (q_4);
    \draw[trans] (q_1) edge node[below right]{\lock} (q_3);
    \draw[trans] (q_1) edge node[below left]{\open} (q_2);
    \draw[trans] (q_4) edge node[below right]{\unlock} (q_2);
    \draw[trans] (q_4) edge[bend left] node[left]{\close} (q_5);
    \draw[trans] (q_5) edge[bend left] node[left]{\open} (q_4);
    \draw[trans] (q_5) edge node[left]{\unlock} (q_1);
    \draw[trans] (q_2) edge[bend right] node[above right]{\close} (q_1);
    \draw[trans] (q_2) edge[bend right] node[below]{\lock} (q_6);
    \draw[trans] (q_6) edge[bend right] node[above]{\unlock} (q_2);
    \draw[trans] (q_6) edge[bend left] node[above left]{\close} (q_7);

    \draw[trans] (q_7) edge[bend left, color=red] node[below left]{\open : 0.1} (q_6);
    \draw[trans] (q_7) edge[bend left, color=red] node[above left]{\open : 0.9} (q_4);
    \draw[trans] (q_7) edge[bend right] node[below left]{\unlock} (q_1);
\end{tikzpicture}\label{fig:Stoc Car}
}

\end{tabular}
        \caption{Car alarm systems with deterministic (\subref{fig:Det Car}) and probabilistic (\subref{fig:Stoc Car}) behavior.}
        \label{fig:Car}
    \end{figure}
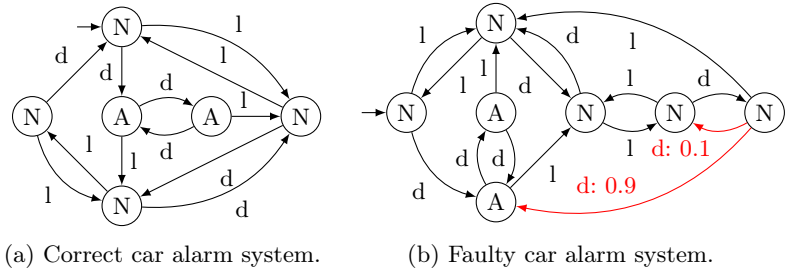
    
    Consider a car which can be locked/unlocked (l) and its doors opened/ closed (d). The car also has an alarm, which triggers when the car is opened in a locked state and can only be disabled by unlocking the car. Figure \ref{fig:Det Car} shows a Moore machine that models such a car alarm system. Depending on its state, its output is either alarm (A) or no alarm (N). Assume we do not know the real model and want to extract a model from recorded execution traces using   
    RPNI \cite{RPNI}, a well-known automata learning algorithm for DFAs which has been extended to Moore and Mealy machines in learning frameworks such as LearnLib \cite{IsbernerHS15} and AALpy. 
    Assuming the variable \code{data} contains known traces from the car alarm system, a Moore machine can be learned using the following code.
    \begin{pythoncode}
    data = [["N", ("d","A"), ("d","A"), ("l","N"), ("d","N")], ... ]
    
    alg = GeneralizedStateMerging(output_behavior="moore",
        transition_behavior="deterministic")
    model = alg.run(data)
    \end{pythoncode}
    No \code{ScoreCalculation} is provided, since RPNI greedily accepts possible merges without scoring and determinism is automatically enforced when specifying deterministic transition behavior. 
    Mealy machines can be learned simply by setting \code{output\_behavior="mealy"}, assuming \code{data} has the correct format.
\end{example}

The \code{ScoreCalculation} class can be instantiated with two parameters, which control local compatibility criteria and scoring. Those parameters are functions operating on the \code{GsmNode} class, which represents automaton states of IOFA.
\begin{description}
    \item \code{local\_compatibility} is a function that takes two \code{GsmNode} objects and returns whether they are locally compatible. The first argument corresponds to the partition created up to that point and the second argument to the state to be added to the partition. Incompatibility of any two tested nodes leads to immediate rejection of the merge candidate.
    \item \code{score\_function} is a function that takes a dictionary mapping states prior to merging to the state describing the behavior of their associated partition. The function should return some form of score. The values \code{True} and \code{False} correspond to a score of positive and negative infinity and leads to immediate acceptance and rejection of the merge respectively.
\end{description}

\begin{example}\label{ex:EDSM}
    The Evidence Driven State Merging (EDSM) algorithm \cite{LangPP98} is similar to RPNI, except that it also uses scoring in order to make better informed decisions. The score of a merge candidate depends on the amount of evidence available supporting the merge. This is quantified as the number of local compatibility checks triggered by merging the two states.
    
    \begin{pythoncode}
    def EDSM_score(part : Dict[GsmNode, GsmNode]):
        nr_partitions = len(set(part.values()))
        nr_merged = len(part)
        return nr_merged - nr_partitions
    
    score = ScoreCalculation(score_function=EDSM_score)
    alg = GeneralizedStateMerging(score_calc=score)
    \end{pythoncode}
\end{example}

An example implementing the IOAlergia algorithm \cite{IOAlergia} for probabilistic systems such as the faulty car alarm depicted in Figure \ref{fig:Stoc Car} using a local compatibility criterion can be found in the appendix (Ex. \ref{ex:IOAlergia}).

The \code{ScoreCalculation} class can also be subclassed to create stateful score functions, which accumulate information from the local compatibility checks as the partitioning is created. In addition to the two methods described above, stateful implementations may override the method \code{reset} which is used to reset the internal state between score calculations. An example is provided in the appendix (Ex. \ref{ex:IOAlergia + EDSM}).

Apart from the options listed above, \code{GeneralizedStateMerging} also has some minor parameters that change details in the red-blue framework. The options \code{pta\_processing} and \code{postprocessing} allow to manipulate the internal representation before and after state merging. They can be set to arbitrary functions mapping one \code{GsmNode} to another and are applied to the initial state. 
Setting the flags \code{eval\_compat\_on\_pta} and \code{eval\_compat\_on\_futures} changes the way how local compatibility is evaluated, which is required e.g. to implement IOAlergia. Using the former causes the algorithm to use the original PTA for evaluating compatibility rather than the current automaton. The latter restricts compatibility checks to common futures, rather than computing it with respect to the partition implied by the merge candidate.
The parameter \code{node\_order} allows to specify the order in which merge candidates are evaluated. This is especially relevant for algorithms which greedily accept the first feasible merge without the use of scores. The default is the shortlex order of the prefix.
The flag \code{consider\_only\_min\_blue} further restricts the set of merge candidates to pairs of red states with the minimal blue node according to the node order.
Finally, setting the flag \code{depth\_first} causes the algorithm to search for violations of the local compatibility criterion in a depth first manner rather than breath first.

\subsection{Utilities}

The \code{run} method of the \code{GeneralizedStateMerging} class also has optional parameters that facilitate the analysis of algorithms. They can be used when experimenting with new algorithms or to better understand under which circumstances existing algorithms fail. 
\begin{description}
    \item The parameter \code{convert} can be set to \code{False} to avoid converting the internal representation to the specific automaton type and causes \code{run} to return a \code{GsmNode} object.
    \item The parameter \code{instrumentation} can be used to instrument the red-blue framework using an \code{Instrumentation} object that provides the option for callbacks at important points (merging, promotion, etc.).
\end{description}
Apart from the input format of input-output traces, other input formats are supported depending on the setting. In DFA inference, the training data is usually not prefix-closed. In this case, the input data can be represented as a set of labeled input sequences, where each element is a tuple consisting of an input sequence and a single output corresponding to the observation after applying the input sequence. For probabilistic models without input-output behavior, we support data specified as sequences of observations only. While the data format is usually detected automatically, it can be explicitly specified using the \code{data\_format} argument of the \code{run} method.

The \code{visualize} method of the \code{GsmNode} class provides highly customizable visualization of automata based on the dot format of GraphViz \cite{EllsonGKNW00}. For example, this could be used to highlight specific aspects of a model such as states that are locally deterministic, or to visualize incorrect merges during the design phase of new algorithms for debugging purposes.

\section{Concluding Remarks}
\label{sec:conc}

There are several software packages for automata learning. 
LearnLib \cite{IsbernerHS15} is very similar to AALpy in general. It mainly provides active automata learning for deterministic IO automata with a few passive algorithms. While its focus is more narrow, LearnLib provides more algorithms in its domain. The same holds true for jajapy \cite{ReynouardIB23}, which is mainly concerned with passive learning of probabilistic models. The library libalf \cite{BolligKKLNP10} falls in a similar category, but seems to be no longer actively developed.
FlexFringe \cite{VerwerH22} is a C++ implementation of the red-blue framework and thus closely related to the presented extension of AALpy. A major difference is that FlexFringe does not support IO behavior out of the box. Supporting IO behavior requires a workaround, especially in the case of observably nondeterministic systems. On the other hand, FlexFringe offers more configuration options. A particularly noteworthy feature is the option to attach custom data to states which is exposed to the compatibility and scoring functions.
The tools RALib \cite{Cassel2015RALibA}, Tomte \cite{AartsHKOV12} and Mint \cite{WalkinshawTD16} all deal with learning of Extended Finite State Machines in some form, which is not covered in AALpy. 

We extended the passive automata learning capabilities of AALpy by adding a general implementation of the red-blue framework \cite{LangPP98}. This implementation is general enough to cover state-merging algorithms for various types of IO automata which may exhibit deterministic, (observably) nondeterministic or probabilistic transition behavior and Mealy- or Moore-style output behavior. While the main focus was to provide a method that is general yet simple to use rather than performance, the provided implementation is sufficiently fast for practical applications. For example, we learned DFAs with well over thousand states from several hundred thousand traces in only a few minutes using the \code{pypy} interpreter on an ordinary laptop. Still, when comparing the RPNI implementations of LearnLib and our framework using the training data from the Abbadingo competition \cite{LangPP98}, our implementation is slower than LearnLib by a factor of 7.24. For the more complex (i.e. time consuming) Abbadingo tasks, the difference is less pronounced with a factor of 6.3. This constitutes a significant but not prohibitive difference. We assume the main causes for the difference to be (1) the lower performance of Python compared to Java and (2) our more general implementation supporting nondeterministic transitions, which is not required for RPNI and thus overhead. If performance is an issue, a dedicated implementation in a low level language is of course preferable. This is not our intention for AALpy and this extension.
In some cases, the implementation abstracts the algorithm based on the parameters to reduce the overhead due to the generality of the red-blue framework.
All in all, this results in a versatile state-merging tool which we expect to be useful to both researchers and industry.

In the future, we want to us the new capabilities of AALpy to explore combinations of active and passive algorithms. This can be particularly interesting when domain knowledge is available that can be used to efficiently generate traces with interesting (e.g. safety critical) behavior. We also aim to use the provided framework to explore how domain knowledge about systems can be injected into existing algorithms, which can be helpful in low data scenarios \cite{BergARST24}.

\textbf{Acknowledgements.} 
This work has received funding from the AIDOaRt project (grant agreement No 101007350) from the ECSEL Joint Undertaking (JU). The JU receives support from the European Union’s Horizon 2020 research and innovation programs and Sweden, Austria, Czech Republic, Finland, France, Italy, and Spain.

The research leading to these results has received funding from the Transformative AI-Assisted Testing in Industrial Mobile Robotics (TASTE) project (project number FO999911053) from the ICT of the Future program. ICT of the Future is a research, technology and innovation funding program of the Republic of Austria, Ministry of Climate Action.

\textbf{Disclosure of Interests.} The authors have no competing interests to declare that are relevant to the content of this article.

\appendix 
\section*{Appendix}\label{appendix}
This appendix provides a collection of examples which illustrate how state-merging algorithms can be defined and executed in our tool. The examples mostly cover algorithms for learning probabilistic models. At the end, we compare the implemented algorithms in a low-data scenario. The resulting automata are shown in Figure \ref{fig:compare}.

\begin{example}\label{ex:IOAlergia}
    Consider the car alarm system from Example \ref{ex:RPNI}. A faulty car alarm system might fail to register a closing door when the car is locked but open with a failure rate of 10\%. A probabilistic model of the faulty system is shown in Figure \ref{fig:Stoc Car} with the probabilistic transitions marked in red. Traces from this system may show nondeterministic behavior, which causes RPNI to fail. We can learn a probabilistic model using IOAlergia \cite{IOAlergia}, an algorithm for learning automata with Moore-style output behavior and probabilistic transition behavior. For each input which is enabled for the two states in consideration, the compatibility function of IOAlergia checks whether the respective distributions of outputs are similar enough, considering the amount of available data. This check is based on the Hoeffding-bound, for which we will use the stub \code{hoeffding\_compat} in the code. The method has a single parameter $\epsilon$, which determines, how aggressively the algorithm merges states.
    The algorithm only evaluates the compatibility on the shared futures of the merge candidate and not on the resulting partitioning. For statistical reasons it also does not evaluate the compatibility on the intermediate hypothesis, but on the PTA. For each transition, the observation counts of the corresponding transition in the PTA can be accessed using the field \code{original\_count}.

    \begin{pythoncode}
    def ioalergia_compat(a: GsmNode, b: GsmNode):
        transition_dummy = TransitionInfo(None, 0, None, 0)
        # iterate over inputs that are common to both states
        in_iter = intersection_iterator(a.transitions, b.transitions)
        for in_sym, a_trans, b_trans in in_iter:
            a_tot = sum(x.original_count for x in a_trans.values())
            b_tot = sum(x.original_count for x in b_trans.values())
            # iterate over outputs that appear in either distribution
            out_iter = union_iterator(a_trans, b_trans, transition_dummy)
            for out_sym, a_trans, b_trans in out_iter:
                a_cnt = a_trans.original_count
                b_cnt = b_trans.original_count
                if not hoeffding_compat(a_cnt, a_tot, b_cnt, b_tot, |$\epsilon$|):
                    return False
        return True
        
    alg = GeneralizedStateMerging(transition_behavior="stochastic", 
        compatibility_on_pta=True, compatibility_on_futures=True, 
        score_calc=ScoreCalculation(ioalergia_compat))
    \end{pythoncode}
\end{example}

\begin{example}\label{ex:IOAlergia on Partition}
    In this example, we create a modified version of IOAlergia (see Ex. \ref{ex:IOAlergia}). In this version, merge candidates are evaluated by determining local compatibility between each state $q$ prior to merging and the state corresponding to the partition associated with $q$ after merging.\footnote{Note that this invalidates the assumption of the statistical test because samples are not independent anymore.} This transformation of a local compatibility function to a score function is implemented in the helper function \code{local\_to\_global\_compatibility}. Additionally, \code{original\_count} needs to be exchanged with \code{count} in \code{ioalergia\_compat} to account for the actual frequencies in the current model. 
    
    \begin{pythoncode}
    compat = local_to_global_compatibility(ioalergia_compat)
    alg = GeneralizedStateMerging(transition_behavior="stochastic",
        score_calc=ScoreCalculation(score_function=compat))
    \end{pythoncode}
\end{example}

\begin{example}\label{ex:IOAlergia + EDSM}
    As EDSM scoring is known to be a good heuristic for passive learning, it is reasonable to also apply it in the probabilistic case, e.g. on top of IOAlergia. In most cases, reusing scores is very easy in our framework. However, combining IOAlergia and EDSM as defined above overestimates the evidence. This is because the score function is always evaluated on the resulting partitioning whereas IOAlergia only evaluates common futures on the PTA. However, a stateful \code{ScoreCalculation} object can keep track of the number of local compatibility evaluations as shown below. Assuming a version of \code{ioalergia\_compat} (see Ex. \ref{ex:IOAlergia}) with an additional parameter for $\epsilon$, a dedicated \code{ScoreCalculation} class also makes instantiating algorithms with different parameters more natural for programmers not used to partial application of functions.

    \begin{pythoncode}
    class IOAlergiaWithEDSM(ScoreCalculation):
        def __init__(self, epsilon):
            super().__init__()
            self.epsilon = epsilon
            self.evidence = 0
        
        def reset(self):
            self.evidence = 0
        
        def local_compatibility(self, a: GsmNode, b: GsmNode):
            self.evidence += 1
            return ioalergia_compat(a, b, self.epsilon)
        
        def score_function(self, part: dict[GsmNode, GsmNode]):
            return self.evidence

    score = IOAlergiaWithEDSM(|$\epsilon$|)
    \end{pythoncode}
\end{example}

\begin{example}\label{ex:domain knowledge car alarm}
    In this example, we show how domain knowledge can be incorporated into learning algorithm. Considering the car alarm system, we know that there is a underlying physical state of the car, which can be any combination of locked / unlocked and open / closed. This state can be observed from the inputs by checking the parity of lock actions and door actions respectively. Since all configurations of the physical state influence the behavior of the car alarm system, we can assume that each state in the learned model relates to exactly one physical state. In other words, states with different physical states should not be merged. This can be formalized as a local compatibility criterion and combined with the compatibility criterion of IOAlergia in a straight-forward manner. 

    \begin{pythoncode}
    def car_alarm_parity_compat(a: GsmNode, b: GsmNode):
        def get_parity(node: GsmNode):
            pref = node.get_prefix()
            parity = []
            for key in ["l", "d"]:
                val = sum(in_s == key for in_s, out_s in pref) % 2
                parity.append(val)
            return parity
        return get_parity(a) == get_parity(b)

    def local_compat(a: GsmNode, b: GsmNode, epsilon):
        parity = car_alarm_parity_compat(a, b)
        ioa = ioalergia_compat(a, b, epsilon)
        return parity and ioa
    \end{pythoncode}    
\end{example}

\begin{example}\label{ex:RPNI + Noise}
    In this example we sketch an algorithm for learning deterministic systems in the presence of noisy (i.e. mislabeled) data. We assume knowledge of an estimate of the per-step mislabeling rate \code{error\_rate} and uniformly distributed noise. The quality of a merge candidate can be assessed by calculating how unlikely the resulting partitioning is. This is done by looping over all modified states and inputs while keeping track of the number of observations that do not match the most frequent observation as well as the total number of observations. From those two numbers and the per-step mislabeling probability, the probability $p_{obs}$ of observing at least as many mismatches is calculated using a binomial distribution. The merge is rejected by returning \code{False} if the resulting probability is below a predefined \code{threshold}. Note that this procedure is a non-local compatibility criterion and is thus implemented as a scoring function. The probability $p_{obs}$ is used as score, favoring merges with fewer inconsistencies. This could be extended by also considering some form of evidence similar to EDSM.
    
    \begin{pythoncode}
    from scipy.stats import binom
    def get_main_output(trans: Dict[Any, TransitionInfo]):
        return max(trans.items(), key=lambda x: x[1].count)[0]
        
    def nd_score(part: Dict[GsmNode, GsmNode], error_rate, threshold):
        mismatches = 0
        total_count = 0
        for node in set(part.values()):
            for in_sym, trans in node.transitions.items():
                main_out = get_main_output(trans)
                for out_sym, info in trans.items():
                    if out_sym != main_out:
                        mismatches += info.count
                    total_count += info.count
    
        prob = 1-binom.cdf(mismatches-1, total_count, error_rate)
        if prob < threshold:
            return False
        return prob
    \end{pythoncode}

    A post-processing step is used for removing non-dominant transitions, resulting in a deterministic automaton. However, specifying the transition behavior as deterministic would enforce determinism during runtime, which would reduce the algorithm to RPNI. Thus, we manually convert the resulting model to a deterministic automaton formalism. 
    
    \begin{pythoncode}
    def postprocess(root: GsmNode):
        for node in root.get_all_nodes():
            for in_sym, trans in node.transitions.items():
                output = get_main_output(trans)
                node.transitions[in_sym] = {output: trans[output]}
        return root

    output_behavior = "moore"
    score = lambda part: nd_score(part, error_rate, threshold)
    score = ScoreCalculation(score_function=score)
    alg = GeneralizedStateMerging(transition_behavior="nondeterministic",
        score_calc=score, postprocessing=postprocess)
    internal_model = alg.run(data, convert=False)
    model = internal_model.to_automaton(output_behavior, "deterministic")
    \end{pythoncode}
\end{example}

\subsection*{Comparison of Examples}

Given enough data, the algorithms from Example \ref{ex:IOAlergia}, \ref{ex:IOAlergia on Partition}, \ref{ex:IOAlergia + EDSM} and \ref{ex:domain knowledge car alarm} can learn models of the faulty car alarm system which have the correct structure and correct transition probabilities up to sampling error. To give a rough comparison regarding sample efficiency, we run the examples on a limited data set. In particular, we use 200 samples with lengths chosen uniformly from the range $[10,20]$. The applied inputs are also chosen randomly. For Alergia-based algorithms, we use the default value for $\epsilon$, which is 0.05. We also show the results for the algorithm from Example \ref{ex:RPNI + Noise}. For this algorithm, we roughly estimate the error frequency across all states to be 1\% and use a 5\% significance threshold. Figure \ref{fig:compare} shows the learned models. The result of IOAlergia, shown in Figure \ref{fig:cmp-IOA}, is off by only one state. However, it has been merged too aggressively, which is a known problem in low data scenarios. Unsurprisingly, using the algorithm from Example \ref{ex:IOAlergia on Partition} does not improve the result (see Fig. \ref{fig:cmp-IOA part}). 
While the result achieved by Example \ref{ex:IOAlergia + EDSM} (see Fig. \ref{fig:cmp-IOA EDSM}) better approximates the true structure, it still does not reflect the abstract state space formed by the combination of the locking status and door status. This is evident from the self-transition in the initial state. However, combining IOAlergia with both EDSM and evaluation on the partition results in a model, shown in Figure \ref{fig:cmp-IOA EDSM part}, that contains only a single mistake: the faulty armed state has been merged with the corresponding state in which the system works correctly. The model therefore contains less information on how the fault can be reproduced. The algorithm based on domain knowledge (Ex. \ref{ex:domain knowledge car alarm}) yields the same result (see Fig. \ref{fig:cmp-IOA knowledge}). The algorithm for learning deterministic models in the presence of noise correctly learns the model of the correct car alarm system (see Fig. \ref{fig:cmp-RPNI Noise}).

\newcommand{\compWidth}[0]{0.48}
\begin{figure}
    \centering
    \begin{subfigure}{\compWidth\textwidth}
        \includegraphics[width=0.9\linewidth]{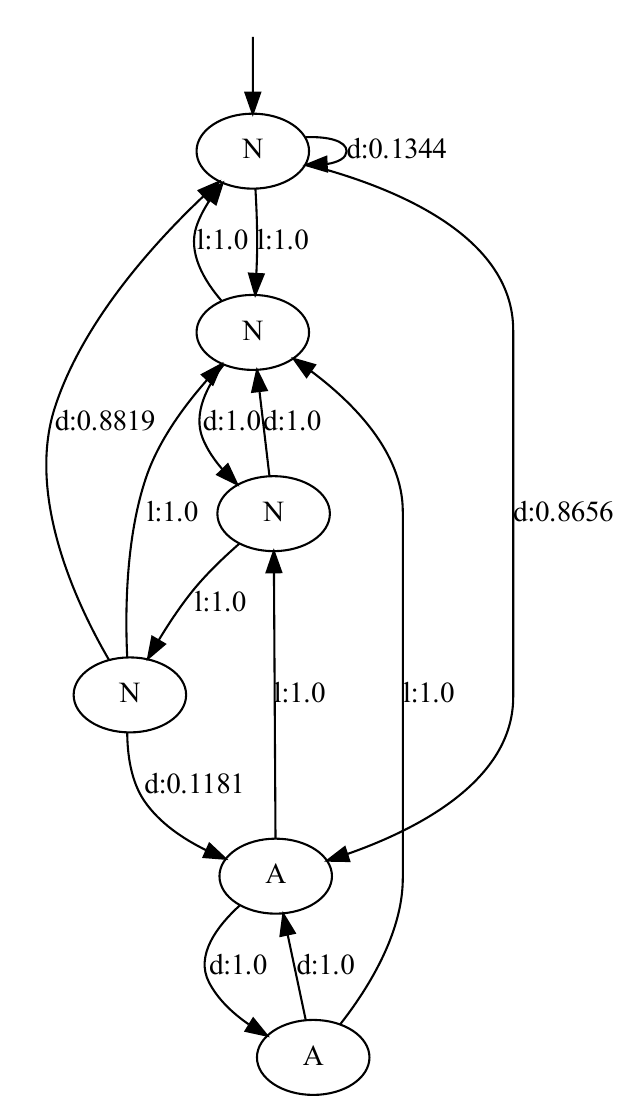}
        \caption{IOAlergia}
        \label{fig:cmp-IOA}
    \end{subfigure}
    \begin{subfigure}{\compWidth\textwidth}
        \includegraphics[width=\linewidth]{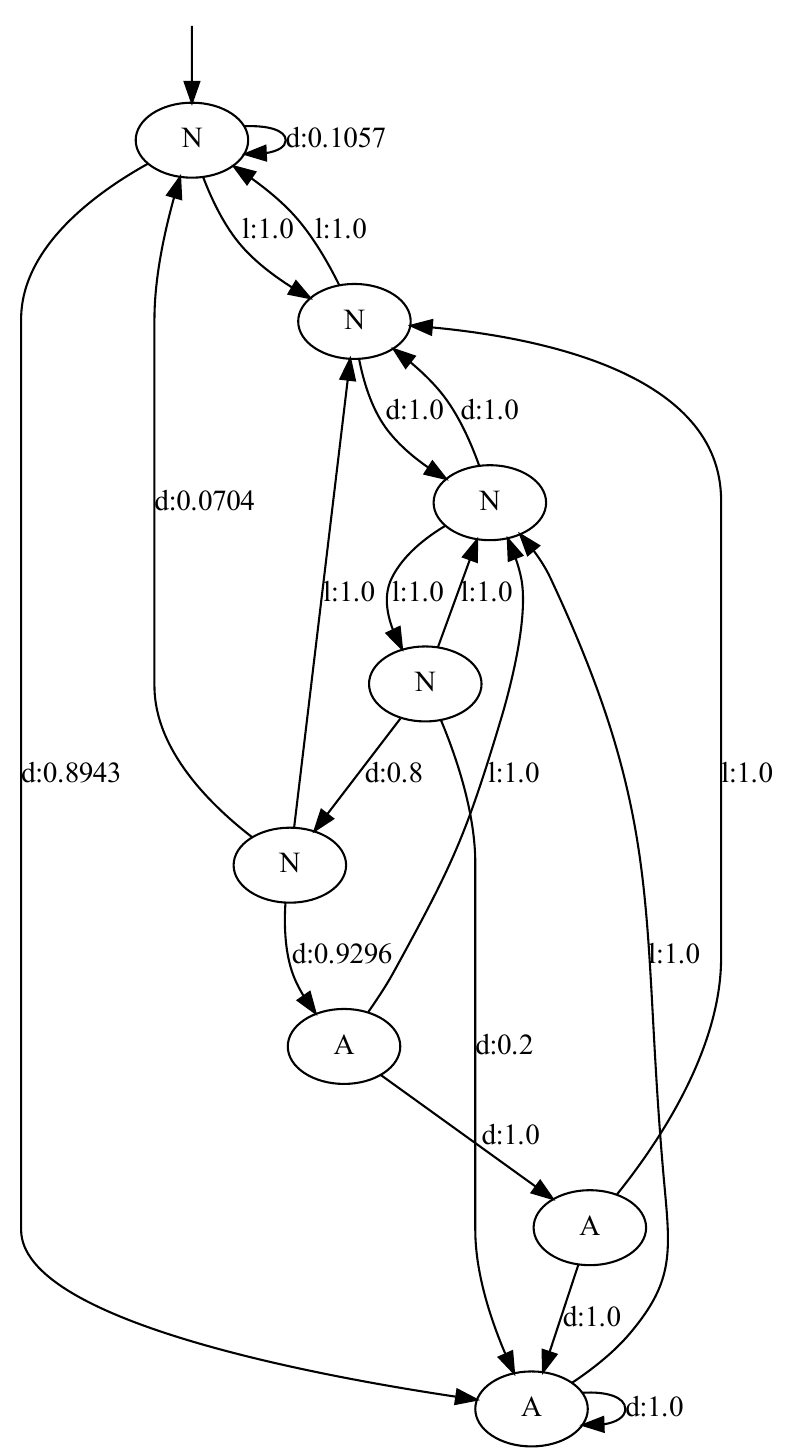}
        \caption{IOAlergia on partition}
        \label{fig:cmp-IOA part}
    \end{subfigure} 
    \caption{Learned models of the faulty car alarm system using different algorithms}
\end{figure}
\begin{figure}
    \ContinuedFloat
    
    \begin{subfigure}{\compWidth\textwidth}
        \includegraphics[width=0.85\linewidth]{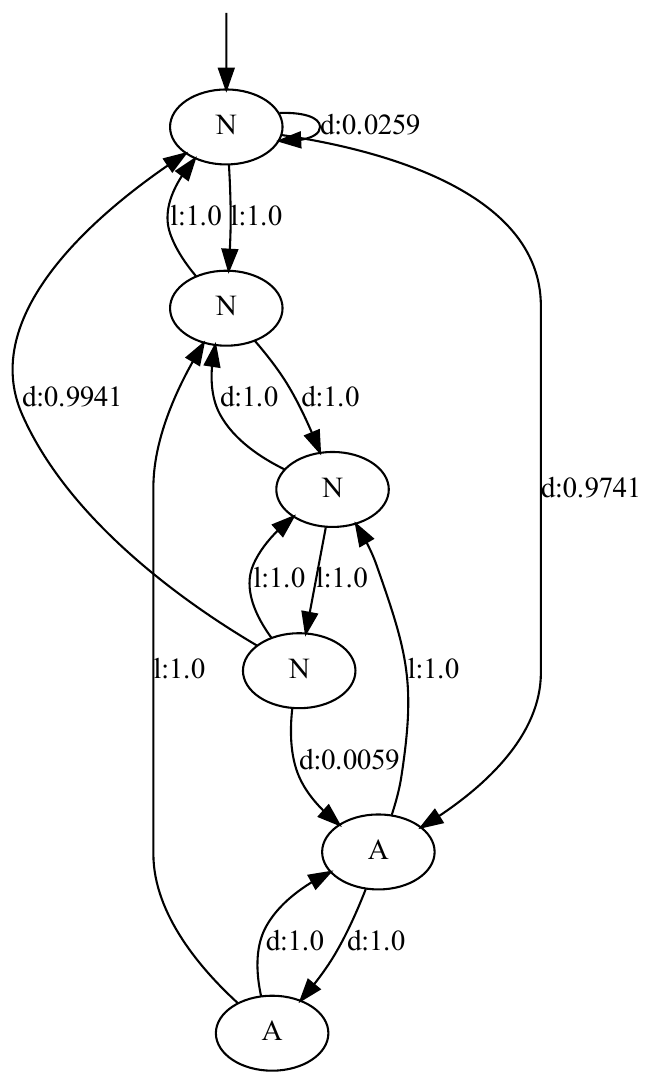}
        \caption{IOAlergia with EDSM}
        \label{fig:cmp-IOA EDSM}
    \end{subfigure}
    \begin{subfigure}{\compWidth\textwidth}
        \centering
        \includegraphics[width=0.85\linewidth]{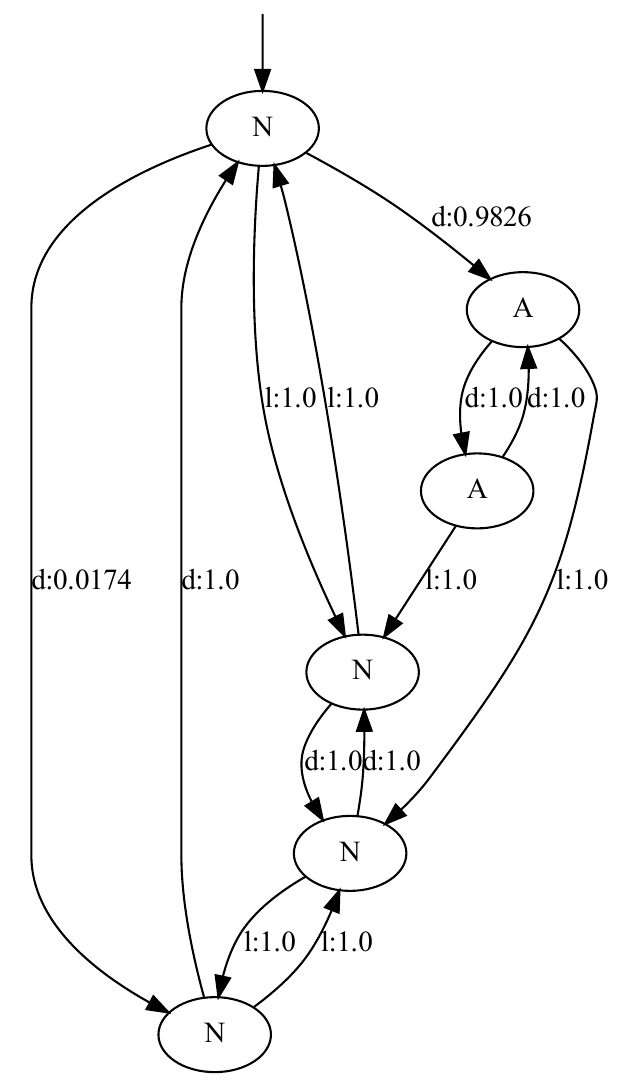}
        \caption{IOAlergia with EDSM on partition}
        \label{fig:cmp-IOA EDSM part}
    \end{subfigure} 
    
    \begin{subfigure}{\compWidth\textwidth}
        \centering
        \includegraphics[width=0.85\linewidth]{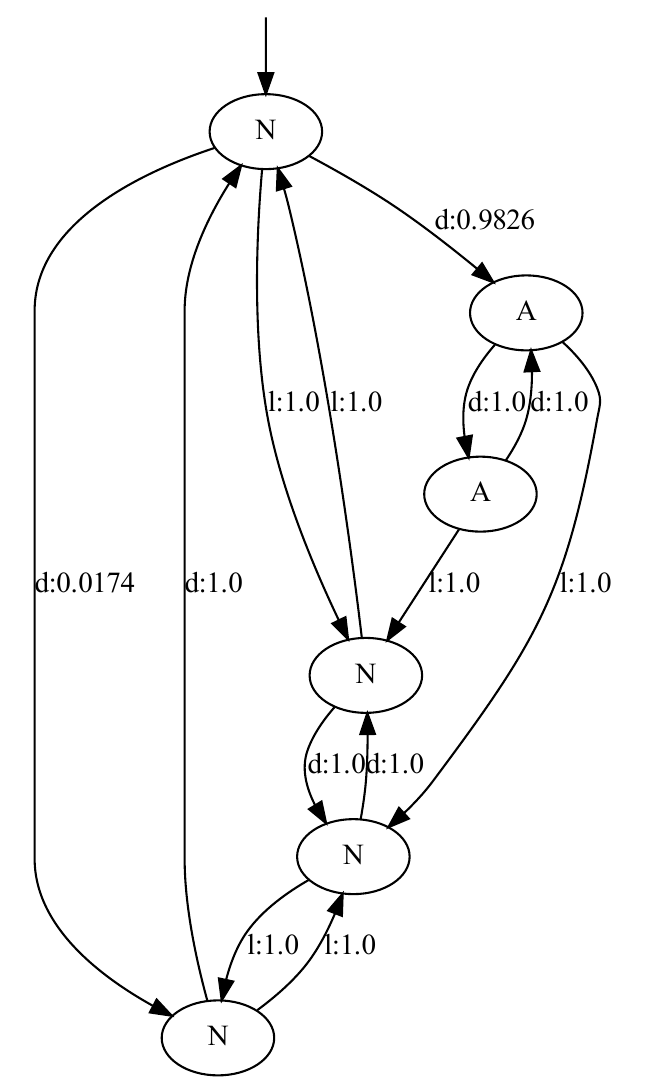}
        \caption{IOAlergia with domain knowledge}
        \label{fig:cmp-IOA knowledge}
    \end{subfigure}
    \begin{subfigure}{\compWidth\textwidth}
        \centering
        \includegraphics[width=0.55\linewidth]{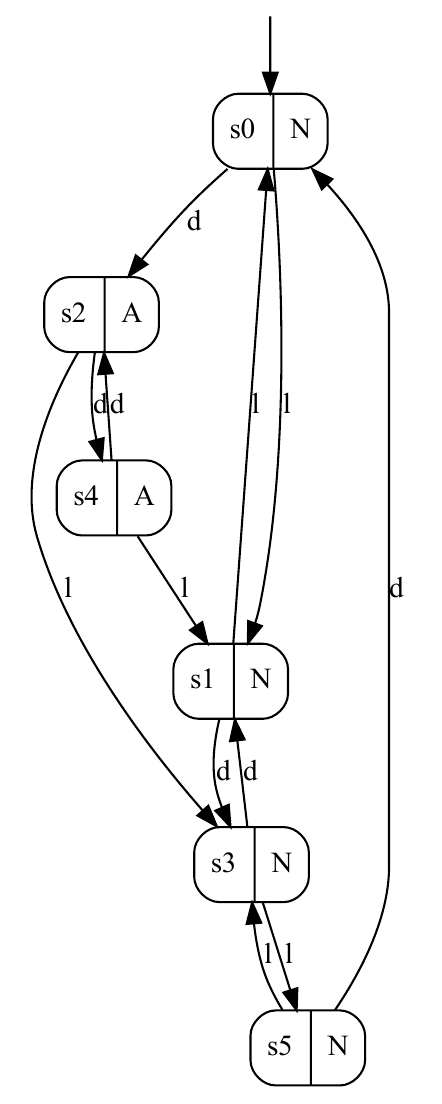}
        \caption{Deterministic learning with noise}
        \label{fig:cmp-RPNI Noise}
    \end{subfigure}
    \caption{Learned models of the faulty car alarm system using different algorithms}
    \label{fig:compare}
\end{figure}

\newpage

\bibliographystyle{splncs04}
\bibliography{main}

\end{document}